\documentclass[runningheads]{llncs}

\usepackage{xspace}
\usepackage{graphicx}
\usepackage{epsfig}
\usepackage{graphicx}
\usepackage{amsmath}
\usepackage{amssymb}
\usepackage{acro}
\usepackage{multirow}
\usepackage{booktabs}
\usepackage{makecell}
\makeatletter
\DeclareRobustCommand\onedot{\futurelet\@let@token\@onedot}
\def\@onedot{\ifx\@let@token.\else.\null\fi\xspace}

\def\eg{\emph{e.g}\onedot} 
\def\ie{\emph{i.e}\onedot}

\def\etal{\emph{et al}\onedot}
\makeatother

\usepackage[pagebackref=true,breaklinks=true,colorlinks,bookmarks=false]{hyperref}

\DeclareAcronym{CAD}{
short=CAD,
long=computer-aided diagnosis,
foreign-plural={}
}

\DeclareAcronym{CXR}{
short=CXR,
long=chest X-ray,
foreign-plural={}
}

\DeclareAcronym{PACS}{
short=PACS,
long=picture archiving and communication system
}

\DeclareAcronym{FROC}{
short=FROC,
long=free-response receiver operating characteristic,
foreign-plural={}
}

\DeclareAcronym{AUROC}{
short=AUROC,
long=area under receiver operating characteristic curve
}

\DeclareAcronym{MIL}{
short=MIL,
long=multi-instance learning,
foreign-plural={}
}

\DeclareAcronym{NLP}{
short=NLP,
long=natural language processing,
foreign-plural={}
}

\DeclareAcronym{FPN}{
short=FPN,
long=feature pyramid network
}

\DeclareAcronym{GT}{
short=GT,
long=ground-truth
}

\DeclareAcronym{EMA}{
short=EMA,
long=exponential moving average
}

\DeclareAcronym{GAP}{
short=GAP,
long=Global Average Pooling
}

\DeclareAcronym{FC}{
short=FC,
long=fully-connected
}

\DeclareAcronym{CAM}{
short=CAM,
long=class activation map
}

\DeclareAcronym{BCE}{
short=BCE,
long=binary cross-entropy
}

\DeclareAcronym{SSL}{
short=SSL,
long=semi-supervised learning
}

\DeclareAcronym{AALS}{
short=AALS,
long=adaptive asymmetric label sharpening
}

\DeclareAcronym{CT}{
short=CT,
long=computed tomography
}

\begin{document}
\title{Knowledge Distillation with Adaptive Asymmetric Label Sharpening for Semi-supervised Fracture Detection in Chest X-rays}
%
%

\author{Yirui Wang\inst{1} \and
Kang Zheng\inst{1}  \and
Chi-Tung Chang\inst{2} \and
Xiao-Yun Zhou\inst{1} \and
Zhilin Zheng\inst{3} \and
Lingyun Huang\inst{3} \and
Jing Xiao \inst{3} \and
Le Lu\inst{1}  \and
Chien-Hung Liao\inst{2} \and
Shun Miao\inst{1}}

\authorrunning{Wang et al.}
%
\institute{PAII Inc., Bethesda, MD, USA \and
Chang Gung Memorial Hospital, Linkou, Taiwan, ROC \and
Ping An Technology, Shenzhen, China}

%
%
\maketitle              
\begin{abstract}

Exploiting available medical records to train high performance \ac{CAD} models via the \ac{SSL} setting is emerging to tackle the prohibitively high labor costs involved in large-scale medical image annotations. Despite the extensive attentions received on \ac{SSL}, previous methods failed to 1) account for the low disease prevalence in medical records and 2) utilize the image-level diagnosis indicated from the medical records. Both issues are unique to \ac{SSL} for \ac{CAD} models. In this work, we propose a new knowledge distillation method that effectively exploits large-scale image-level labels extracted from the medical records, augmented with limited expert annotated region-level labels, to train a rib and clavicle fracture \ac{CAD} model for \ac{CXR}.
Our method leverages the teacher-student model paradigm and features a novel \ac{AALS} algorithm to address the label imbalance problem that specially exists in medical domain. Our approach is extensively evaluated on all \ac{CXR} ($N=65{,}845$) from the trauma registry of Chang Gung Memorial Hospital over a period of 9 years (2008-2016), on the most common rib and clavicle fractures. The experiment results demonstrate that our method achieves the state-of-the-art fracture detection performance, \ie, an \ac{AUROC} of 0.9318 and a \ac{FROC} score of 0.8914 on the rib fractures, significantly outperforming previous approaches by an \ac{AUROC} gap of $1.63\%$ and an \ac{FROC} improvement by $3.74\%$. Consistent performance gains are also observed for clavicle fracture detection.

\end{abstract}

\acresetall

\section{Introduction}

\Ac{CAD} of medical images has been extensively studied in the past decade. In recent years, substantial progress has been made in developing deep learning-based CAD systems to diagnose a wide range of pathologies, \eg, lung nodule diagnosis in chest \ac{CT} images~\cite{xie2018knowledge}, mass and calcification characterization in mammography~\cite{shen2019deep}, bone fracture detection/classification in radiography~\cite{wang2019weakly}. The state-of-the-art CAD solutions are typically developed based on large-scale expert annotations (\eg, 128,175 labeled images for diabetic retinopathy detection~\cite{gulshan2016development}, 14,021 labeled cases for skin condition diagnosis~\cite{liu2020deep}). However, the labor cost of large-scale annotations in medical area is prohibitively high due to the required medical expertise, which hinders the development of deep learning-based CAD solutions for applications where such large-scale annotations are not yet available.
In this work, we aim to develop a cost-effective \ac{SSL} solution to \textit{train a reliable, robust and accurate fracture detection model for \acf{CXR} using limited expert annotations and abundant clinical diagnosis records.}

While expert annotations are expensive to obtain, medical records can often be efficiently/automatically collected retrospectively at large scale from a hospital's information system. Motivated by the availability of retrospective medical records, a few large-scale X-ray datasets with \ac{NLP} generated image-level labels are collected and publicly released, \eg, ChestXray-14~\cite{wang2017chestx}, CheXpert~\cite{irvin2019chexpert}. Previous works have subsequently investigated weakly-supervised learning to train CAD models using purely image-level labels~\cite{rajpurkar2017chexnet,wang2019weakly}. However, since the image-level labels lack localization supervision, these methods often cannot deliver sufficient diagnosis and localization accuracy for clinical usage~\cite{li2018thoracic}. In addition, the public \ac{CXR} datasets rely only on radiology reports, which are known to have substantial diagnostic errors and inaccuracies~\cite{brady2017error}.

A more promising and practical strategy for training CAD models is to use large-scale image-level labels extracted from the \textit{clinical diagnosis reports} with a small number of expert annotated region-level labels. Different from radiology diagnoses made by the radiologist based on a single image modality, clinical diagnoses are made by the primary doctor considering all sources of information, \eg{}, patient history, symptoms, multiple image modalities. Therefore, clinical diagnoses offer more reliable image-level labels for training CAD models. Previous \ac{SSL} methods (\eg., $\rm{\Pi}$-model~\cite{laine2016temporal}, Mean Teacher~\cite{tarvainen2017mean}, Mix-Match~\cite{berthelot2019mixmatch}) have studied a similar problem setup, \ie{}, training classification/segmentation models using a combination of labeled and unlabeled images. 
However, these general-purpose \ac{SSL} methods assume that no label information is given for the \textit{unlabeled} set. Therefore, they fail to take advantage of the clinical diagnosis reports that are available in our application. In addition, there is an unique data imbalance challenge in training CAD models using clinical diagnoses. In particular, due to the low prevalence of fractures in \acp{CXR}, the image-level diagnostic labels are imbalanced toward more negative (\eg{}, 1:10 ratio). The region-level labeled positives and image-level diagnostic negatives are even more imbalanced (\eg{}, 1:100 ratio). As a result, a specifically-designed \ac{SSL} method is required to fully exploit the clinical diagnoses with imbalanced data distribution to effectively train CAD models.

To bridge this gap, we propose an effective SSL solution for fracture detection in \ac{CXR} that better accounts for the imbalanced data distribution and exploits the image-level labels of the unannotated data. We adopt the teacher-student mechanism, where a teacher model is employed to produce pseudo \acp{GT} on the image-level diagnostic positive images for supervising the training of the student model. Different from previous knowledge distillation methods where the pseudo \acp{GT} are directly used or processed with symmetric sharpening/softening, we propose an \textit{\ac{AALS}} to account for the teacher model's low sensitivity caused by the imbalanced data distribution and to encourage positive detection responses on the image-level positive \acp{CXR}.
The proposed method is evaluated on a real-world scenario dataset of all ($N=65{,}843$) \ac{CXR} images taken in the trauma center of Chang Gung Memorial Hospital from year 2008 to 2016. Experiments demonstrate that our method reports an \ac{AUROC} of 0.9318/0.9646 and an \ac{FROC} score of 0.8914/0.9265 on the rib/clavicle fracture detection. Compared to state-of-the-art methods, our method significantly improves the \ac{AUROC} by $1.63\%$/$0.86\%$ and the \ac{FROC} by $3.74\%$/$3.81\%$ on rib/clavicle fracture detection, respectively.

\section{Method}


{\bf Problem Setup} We develop a fracture detection model using a combination of image-level and region-level labeled CXRs. While the image-level labels can be obtained efficiently at a large scale by mining a hospital's image archive and clinical records, the region-level labels are more costly to obtain as they need to be manually annotated by experts. We collected $65{,}845$ \acp{CXR} from the trauma registry of a medical center. Diagnosis code and keyword matching of the clinical records are used to obtain image-level labels, resulting in $6{,}792$ positive and $59{,}051$ negative \ac{CXR}s. Among positive CXRs, $808$ CXRs with positive diagnosis are annotated by experts to provide region-level labels in the form of bounding-box. The sets of region-level labeled, image-level positive and image-level negative CXRs are denoted by $\mathcal{R}$, $\mathcal{P}$ and $\mathcal{N}$, respectively. Our method aims to \textit{effectively exploit both the region-level labels and the image-level labels under extremely imbalanced positive/negative ratio}.


\subsection{Knowledge Distillation Learning}

Similar to recent \ac{CAD} methods~\cite{li2018thoracic}, we train a neural network to produce a probability map that indicates the location of the detected fracture. Since the shape and scale of fractures can vary significantly, we employ \ac{FPN}~\cite{lin2017feature} with a ResNet-50 backbone to tackle the scale variation challenge by fusing multi-scale features. The training consists of two steps: 1) supervised pre-training and 2) semi-supervised training. In the pre-training step, a fracture detection model is trained via supervised learning using $\mathcal{R} \cup \mathcal{N}$. In the semi-supervised training step, $\mathcal{P}$ are further exploited to facilitate the training.

\begin{figure}[t]
	\begin{center}
		\includegraphics[width=\linewidth]{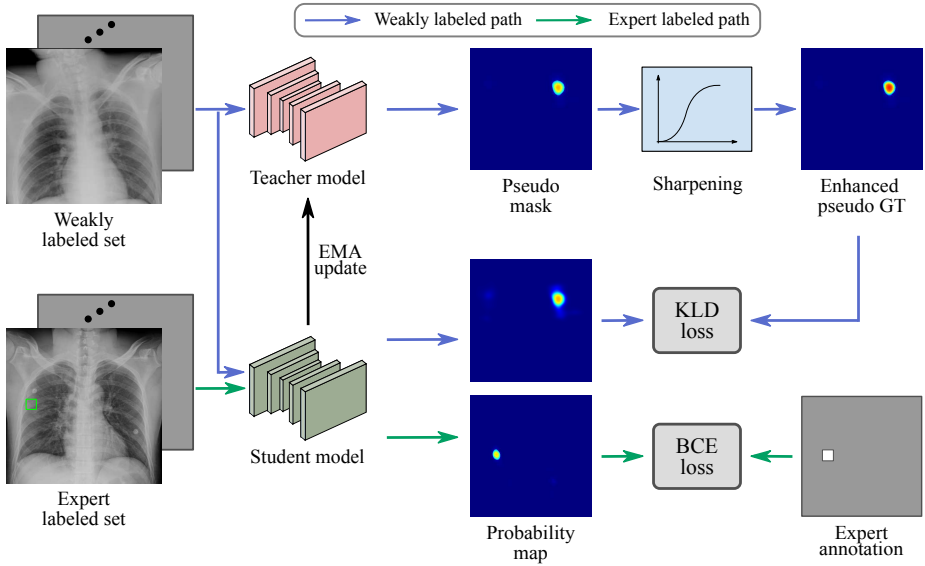}
	\end{center}
	\caption{An overview of the proposed knowledge distillation with \ac{AALS}. The student model is trained via back-propagation. The teacher model is updated by the EMA.}
	\label{fig:system}
\end{figure}

\paragraph{Supervised pre-training}
We train the network using only CXRs in $\mathcal{R}$ and $\mathcal{N}$, where pixel-level supervision signals can be generated. Specifically, for CXRs in $\mathcal{R}$, \ac{GT} masks are generated by assigning \textit{one} to the pixels within the bounding-boxes and \textit{zero} elsewhere. For CXRs in $\mathcal{N}$, \ac{GT} masks with all \textit{zeros} are generated. During training, we use the pixel-wise \ac{BCE} loss between the predicted probability map and the generated \ac{GT} mask, written as:
\begin{equation}
    \mathcal{L}_{sup} = \sum_{x \in (\mathcal{R} \cup \mathcal{N})} \text{BCE} \left(f_\theta(x), y \right),
\end{equation}
where $x$ and $y$ denote the CXR and its pixel-level supervision mask. $f_\theta(x)$ denotes the probability map output of the network parameterized by $\theta$.
Due to the extreme imbalance between $\mathcal{R}$ and $\mathcal{N}$ (\eg{}, 808 vs. 59,861), the pre-trained model tends to have a low detection sensitivity, \ie{}, producing low probabilities on fracture sites. 

\paragraph{Semi-supervised training}
To effectively leverage $\mathcal{P}$ in training, we adopt a teacher-student paradigm where the student model learns from the pseudo \ac{GT} produced by the teacher model on $\mathcal{P}$. The teacher and student models share the same network architecture, \ie{}, ResNet-50 with \ac{FPN}, and are both initialized using the pre-trained weights from the supervised learning step. Inspired by the Mean Teacher method~\cite{tarvainen2017mean}, we train the student model via back propagation and iteratively update the teacher model using the \ac{EMA} of the student model weights during training. Denoting the weights of the teacher and student models at training step $t$ as $\theta^{'}_t$ and $\theta_t$, respectively, the weights of the teacher model are updated following:
\begin{equation}
    \theta^{'}_t = \alpha \theta^{'}_{t-1} + (1-\alpha)\theta_t,
\end{equation}
where $\alpha$ is a smoothing coefficient to control the pace of knowledge update. $\alpha$ is set to 0.999 in all our experiments following~\cite{tarvainen2017mean}.

CXRs in the region-level labeled set ($\mathcal{R}$), image-level labeled positive set ($\mathcal{P}$) and image-level labeled negative set ($\mathcal{N}$) are all used to train the teacher-student model. The training mechanism is illustrated in Fig. \ref{fig:system}. For CXRs in $\mathcal{R}$ and $\mathcal{N}$, the same supervised loss $\mathcal{L}_{sup}$ is used. For CXRs in $\mathcal{P}$, the teacher model is applied to produce a pseudo \ac{GT} map, which is further processed by an \ac{AALS} operator. The sharpened pseudo \ac{GT} of image $x$ is denoted as
\begin{equation}
    y'= S(f_{\theta'_t}(x)),
\end{equation}
where $f_{\theta'_t}$ denotes the teacher model at the $t$-th step, $S(\cdot)$ denotes \ac{AALS}. The KL divergence between the sharpened pseudo \ac{GT} $y'$ and the student model's prediction $f_{\theta_t}(x)$ is calculated as an additional loss:
\begin{equation}
    \mathcal{L}_{semi} = \sum_{x \in \mathcal{P}} \text{KLDiv} \Big(S \left(f_{\theta'_t}(x) \right), f_{\theta_t}(x) \Big).
\end{equation}
The total loss used to train the student network is
\begin{equation}
    \mathcal{L} = \mathcal{L}_{sup} + \mathcal{L}_{semi}.   
\end{equation}

\subsection{Adaptive Asymmetric Label Sharpening}

\begin{figure}[t]
    \begin{minipage}[b]{0.50\linewidth}
        \centering
        \includegraphics[width=\linewidth]{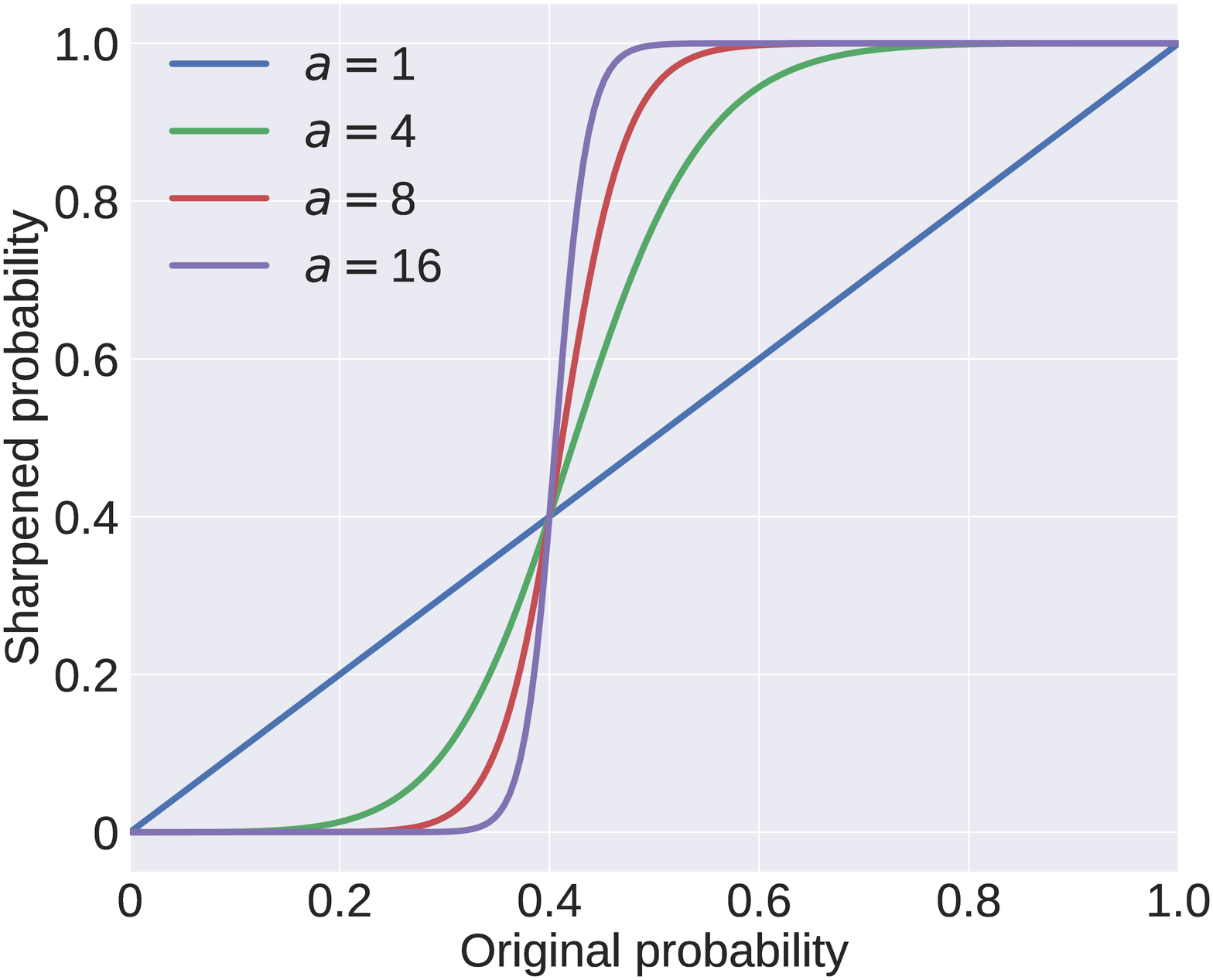}
        \caption{Asymmetric label sharpening function at $t=0.4$ with different $a$.}
        \label{fig:polarization}
    \end{minipage}
    \hfill
    \begin{minipage}[b]{0.50\linewidth}
        \centering
        \includegraphics[width=\linewidth]{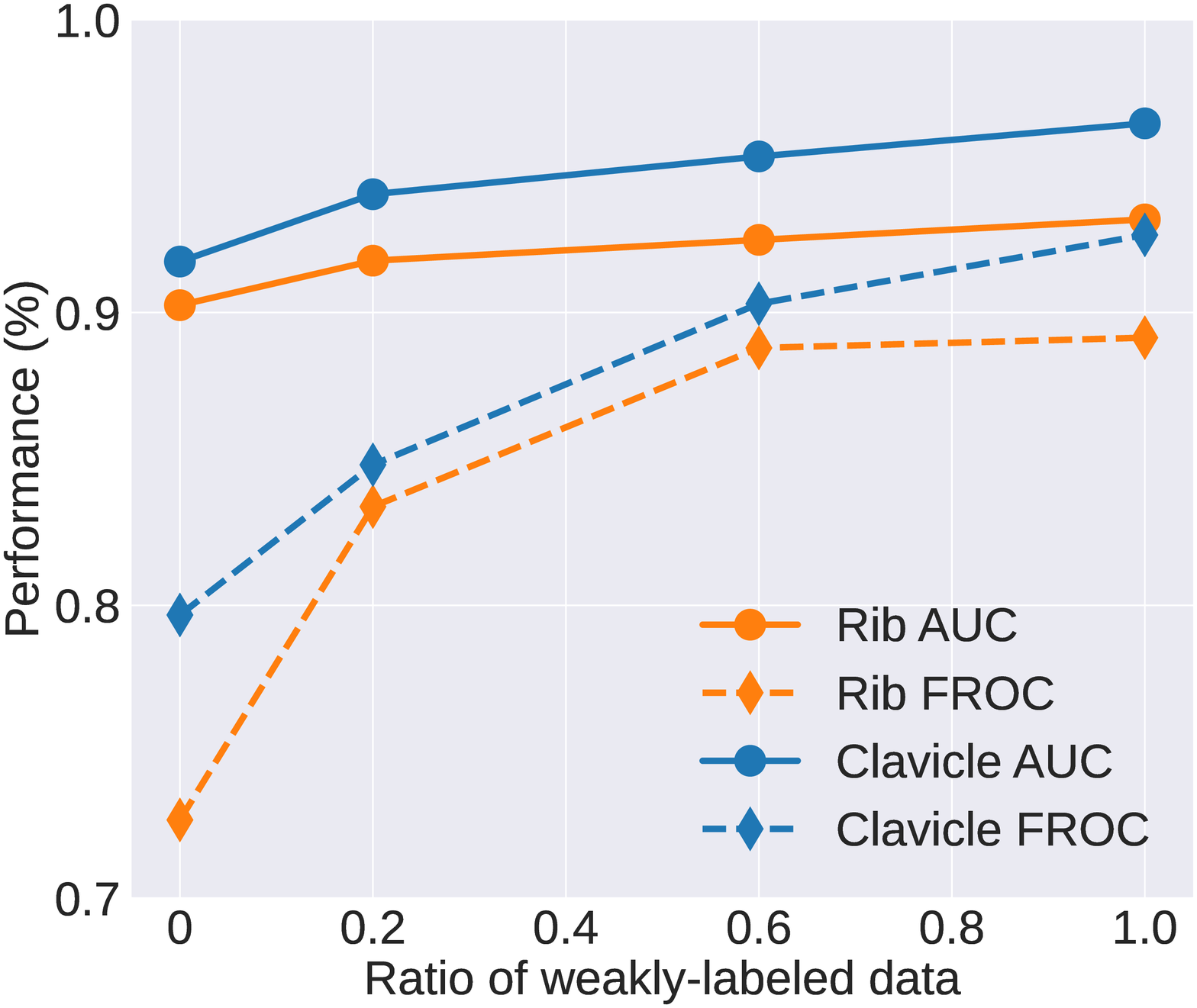}
        \caption{Model performance using a subset of $\mathcal{P}$.}
        \label{fig:data_ratio}
    \end{minipage}
\end{figure}

In previous knowledge distillation models, the pseudo \ac{GT}s are produced on unlabeled data to supervise the student model. Since no prior knowledge is given for the unlabeled data, the pseudo \ac{GT}s are either directly used~\cite{tarvainen2017mean}, or processed with symmetric softening~\cite{hinton2015distilling} or sharpening~\cite{berthelot2019mixmatch}. In our problem setup, we have important prior knowledge that can be exploited: 1) image-level positive CXRs contain visible fracture sites, 2) due to the imbalanced positive/negative ratio, the pseudo \ac{GT} tends to have low sensitivity (\ie{}, low probabilities at fracture sites). Therefore, when the maximum value of the pseudo \ac{GT} map is low, we are motivated to enhance the activation via \ac{AALS}:
\begin{equation}
    S(y') = \text{expit} \big(a \cdot \text{logit}(y') + (1 - a) \cdot \text{logit}(t) \big),
\end{equation}
where $\text{expit}(\cdot)$ and $\text{logit}(\cdot)$ denote Sigmoid function and its inverse. $a$ and $t$ control the strength and center of the sharpening operator, respectively. The effects of $a$ and $t$ are shown in Fig.~\ref{fig:polarization}. Since the asymmetric sharpening aims to enhance the low probabilities in the pseudo \ac{GT}, $t<0.5$ should be used ($t=0.4$ is used in our experiments). Since there are still many fracture sites missed in $y'$ (\ie with low probability values) due to the imbalanced training data, we use $\max(S(y'), y')$ as the label sharpening function in our final solution to avoid over penalization of the student model's activation on fracture sites with low probability values in $y'$.

The sharpening strength $a$ is dynamically selected based on the maximum probability in the pseudo \ac{GT} map, written as:
\begin{equation}
    a = a_0 - (a_0 - 1) y'_{max},
\end{equation}
where $y'_{max}$ is the maximum probability in the pseudo \ac{GT} map, $a_0$ is a hyperparameter that controls the largest sharpening strength allowed. The sharpening strength $a$ is negatively correlated with the maximum probability $y'_{max}$. When $y'_{max}$ approaches 1, $a$ approaches its minimum value 1, making $S(\cdot)$ an identity mapping. When $y'_{max}$ decreases, $a$ increases toward $a_0$, leading to stronger sharpening of the pseudo \ac{GT}. A dynamic $a$ is required because the sharpening operator is asymmetric. If a constant $a>1$ is used, the sharpening operation will always enlarge the activation area in the pseudo \ac{GT} map, which drives the model to produce probability maps with overly large activation areas. With the adaptive sharpening strength, when a fracture site is confidently detected in a \ac{CXR} (\ie, $y'_{max}$ approaches 1), the sharpening operation degenerates to identity mapping to avoid consistently expanding the activation area.

\setlength{\tabcolsep}{3pt}

\subsection{Implementation Details}

We trained our model on a workstation with a single Intel Xeon E5-2650 v4 CPU @ 2.2 GHz, 128 GB RAM, 4 NVIDIA Quadro RTX 8000 GPUs. All methods are implemented in Python 3.6 and PyTorch v1.6. We use the ImageNet pre-trained weights to initialize the backbone network. Adam optimizer is employed in all methods. A learning rate of $4e-5$, a weight decay of $0.0001$ and a batch size of $48$ are used to train the model for $25$ epochs. All images are padded to square and resized to $1024 \times 1024$ for network training and inference. We randomly perform rotation, horizontal flipping, intensity and contrast jittering to augment the training data. The trained model is evaluated on the validation set after every training epoch, and the one with the highest validation \ac{AUROC} is selected as the best model for inference.

\begin{figure}[t]
    \centering
	\includegraphics[width=0.98\linewidth]{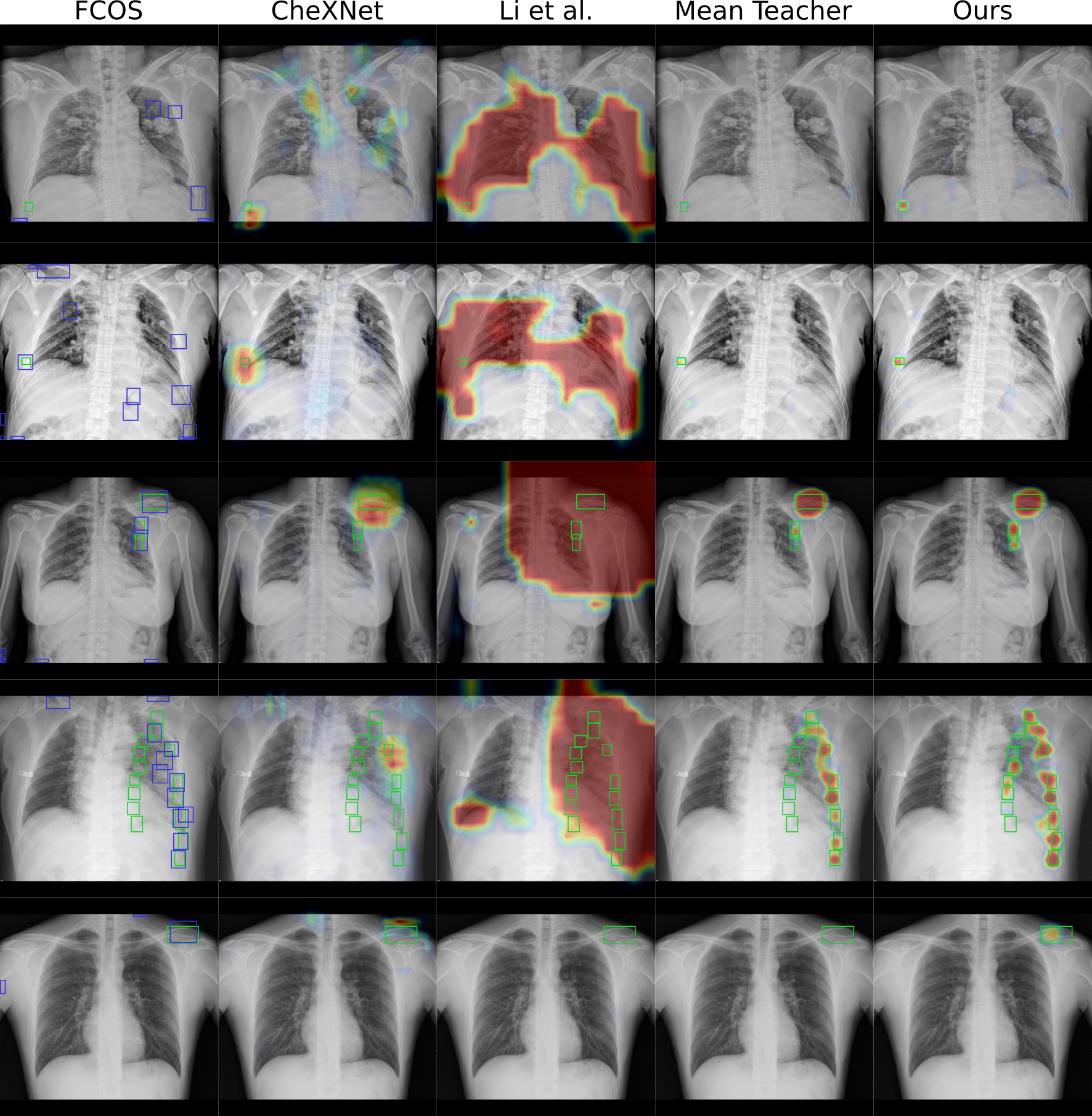}
	\caption{Examples of the fracture detection results. \ac{GT} and FCOS detected fracture bounding-boxes are shown in green and blue colors.}
	\label{fig:fracture_visualization} \vspace{-3mm}
\end{figure}

\section{Experiments}
\subsection{Experimental Settings}

\paragraph{\textbf{Dataset}}
We collected $65{,}843$ CXRs of unique patients that were admitted to the trauma center of Chang Gung Memorial Hospital from year 2008 to 2016. Based on the clinical diagnosis records, the CXRs are assigned image-level labels for rib and clavicle fractures. In total, we obtained $6{,}792$ ($\mathcal{R} \cup \mathcal{P}$) CXRs with positive label for at least one type of fracture and $59{,}051$ ($\mathcal{N}$) CXRs with negative label for both fracture types. $808$ ($\mathcal{R}$) image-level positive CXRs are randomly selected for expert annotation by two experienced trauma surgeons. The annotations are confirmed by the best available information, including the original CXR images, radiologist reports, clinical diagnoses, and advanced imaging modality findings (if available). All experiments are conducted using five-fold cross-validation with a $70\%$/$10\%$/$20\%$ training, validation, and testing split, respectively.

\setlength{\tabcolsep}{4mm}
\begin{table}[t]
\centering
\caption{Fracture classification and localization performance comparison with state-of-the-art models. AUROC is reported for classification performance. FROC score is reported for localization performance.}
\vspace{1em}
\begin{tabular}{lcccc}
\toprule
\multirow{2}{*}{Method}  & \multicolumn{2}{c}{Rib fracture} & \multicolumn{2}{c}{Clavicle fracture} \\ 
                           \cmidrule(lr){2-3}                 \cmidrule(lr){4-5}
                                              & AUROC & FROC                  & AUROC & FROC  \\ \midrule
CheXNet~\cite{rajpurkar2017chexnet}           & 0.8867  & -       & 0.9555  & -\\ \midrule
RetinaNet~\cite{lin2017focal}                 & 0.8609  & 0.4654  & 0.8610  & 0.7985 \\
FCOS~\cite{tian2019fcos}                      & 0.8646  & 0.5684  & 0.8847  & 0.8471 \\ \midrule
Li \etal{}~\cite{li2018thoracic}              & 0.8446  & -       & 0.9560  & - \\
$\rm{\Pi}$-Model~\cite{laine2016temporal}     & 0.8880  & 0.7703  & 0.9193  & 0.8536 \\
Temporal Ensemble~\cite{laine2016temporal}    & 0.8924  & 0.7915  & 0.9132  & 0.8204 \\
Mean Teacher~\cite{tarvainen2017mean}         & 0.9155  & 0.8540  & 0.9474  & 0.8884 \\ \midrule
Supervised pre-training                       & 0.9025  & 0.7267  & 0.9174  & 0.7967 \\
\multirow{2}{*}{\textbf{Ours}}                & \textbf{0.9318}   & \textbf{0.8914}  & \textbf{0.9646}  & \textbf{0.9265} \\
& \scriptsize{\color{blue} (+1.63\%)} & \scriptsize{\color{blue} (+3.74\%)}  & \scriptsize{\color{blue} (+0.86\%)} & \scriptsize{\color{blue} (+3.81\%)}\vspace{1mm} \\
\bottomrule
\end{tabular}
\label{tab:existing_method_comparison}
\end{table}

\paragraph{\textbf{Evaluation metrics}}
We evaluate both fracture classification and localization performances. The widely used classification metric \ac{AUROC} is used to assess classification performance. For object detection methods, the maximum classification score of all predicted bounding-boxes is taken as the classification score. For methods producing probability map, the maximum value of the probability map is taken as the classification score. We also assess the fracture localization performance of different methods. Since our method only produces probability map, standard \ac{FROC} metric based on bounding-box predictions is infeasible. Thus, we report a modified \ac{FROC} metric to evaluate the localization performance of all compared methods. A fracture site is considered as recalled if the center of its bounding-box is activated. And the activated pixels outside bounding-boxes are regarded as false positives. Thus, the modified \ac{FROC} measures the fracture recall and the average ratio of false positive pixels per image. To calculate the modified \ac{FROC} for object detection methods, we convert their predicted bounding-boxes into a binary mask using different thresholds, with the pixels within the predicted box as positive, and the pixels outside the box as negative. To quantify the localization performance, we calculate an FROC score as an average of recalls at ten false positive ratios from 1\% to 10\%.

\paragraph{\textbf{Compared methods}}
We compare the proposed method with baseline methods in the following three categories. 1) \textbf{Weakly-supervised methods:} We evaluate \textit{CheXNet}~\cite{rajpurkar2017chexnet}, a representative state-of-the-art X-ray CAD method trained purely using image-level labels. 2) \textbf{Object detection methods:} We evaluate two state-of-the-art object detection methods: an anchor-based detector \textit{RetinaNet}~\cite{lin2017focal} and an anchor-free detector \textit{FCOS}~\cite{tian2019fcos}. 3) \textbf{Semi-supervised methods: } We evaluate three popular knowledge distillation methods, $\rm{\Pi}$-\textit{Model}~\cite{laine2016temporal}, \textit{Temporal Ensemble}~\cite{laine2016temporal} and \textit{Mean Teacher}~\cite{tarvainen2017mean}, and a state-of-the-art medical image \ac{SSL} method by Li \etal{}~\cite{li2018thoracic}. For all evaluated methods, ResNet-50 is employed as the backbone network. FPN is employed in the two detection methods, RetinaNet and FCOS. 

\subsection{Comparison with Baseline Methods}

\begin{figure}[t]
    \centering
	\includegraphics[width=0.49\linewidth]{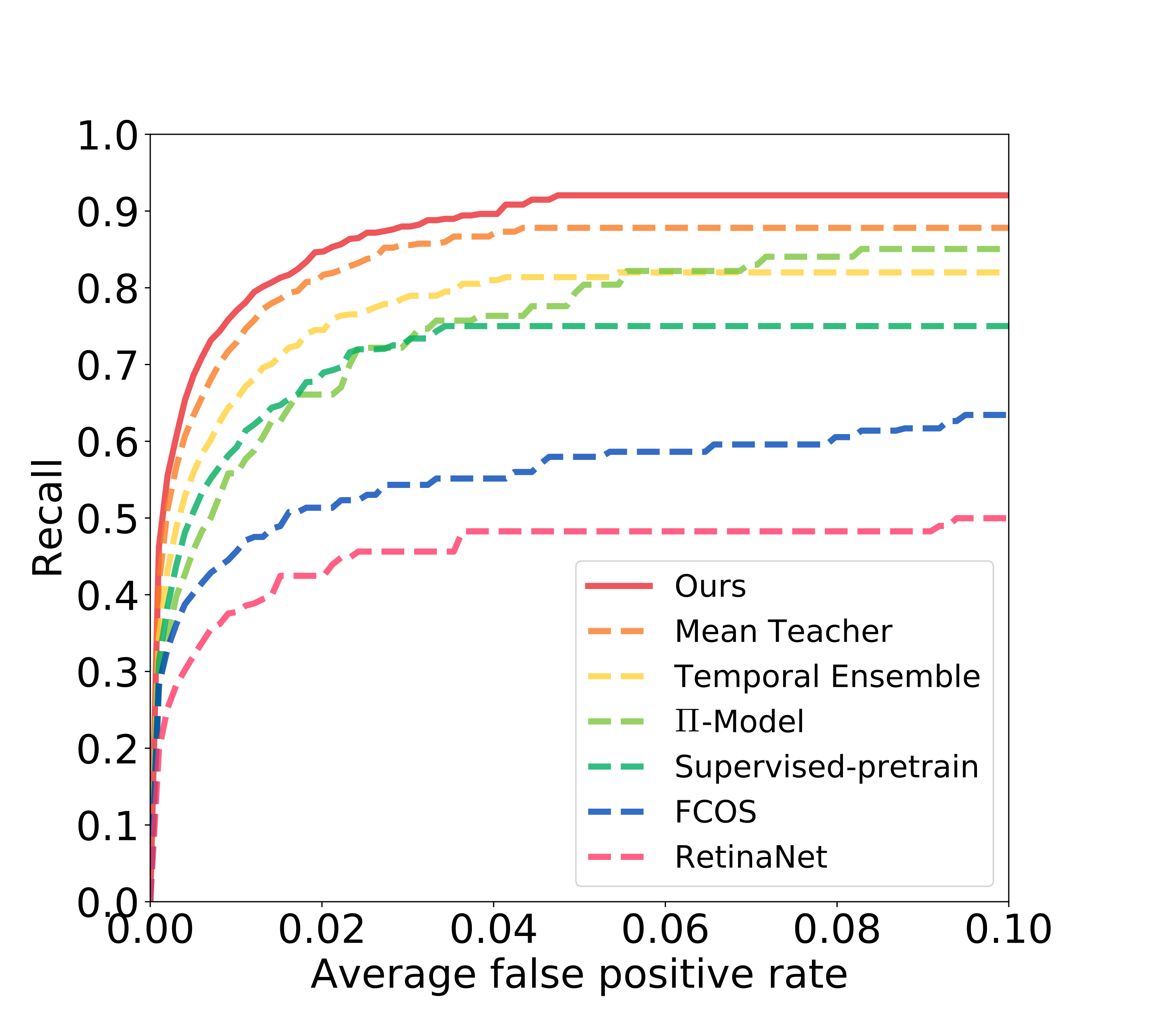}
	\includegraphics[width=0.49\linewidth]{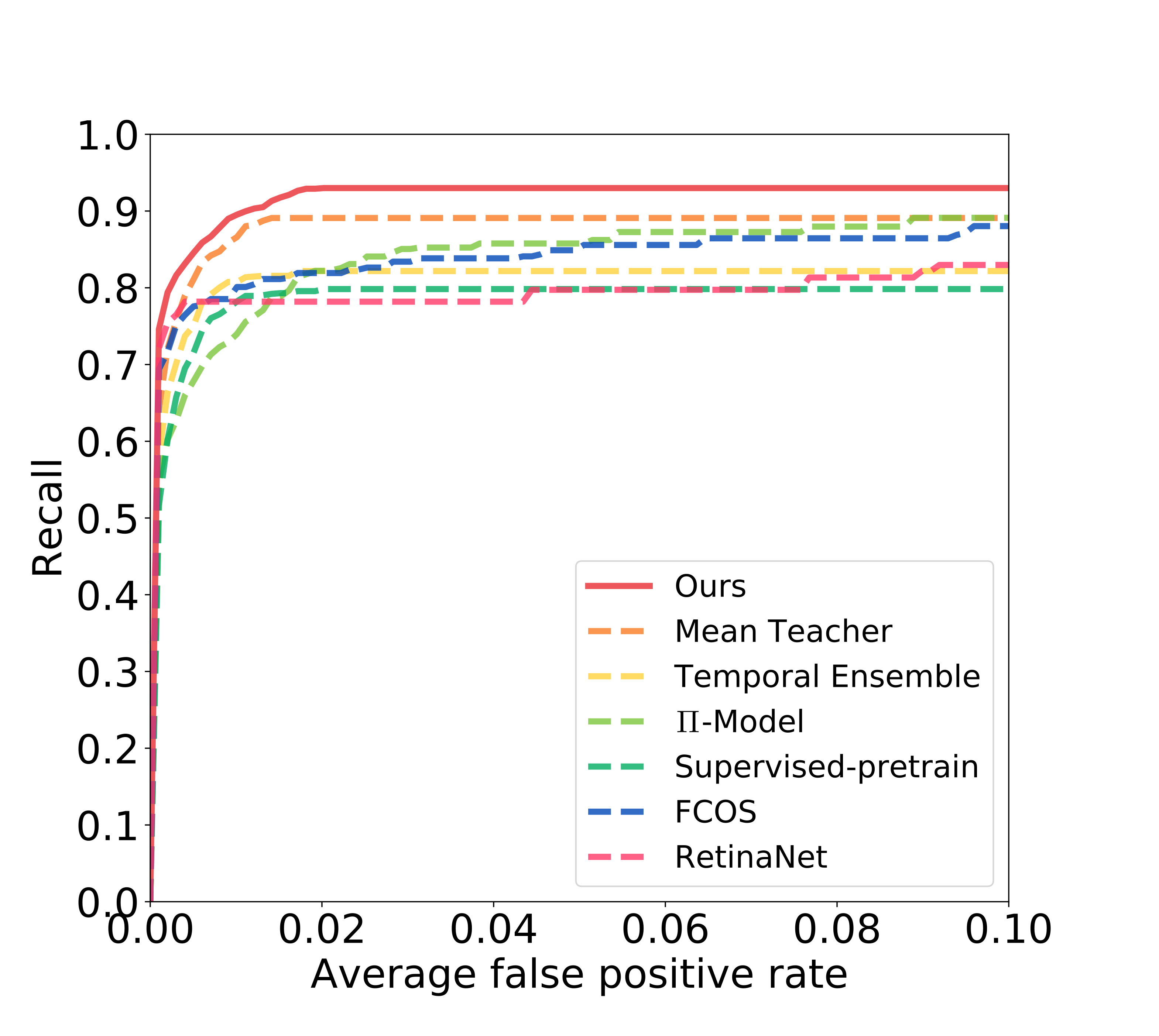}
	\caption{FROC curves of rib fracture (left) and clavicle fracture (right) detection results using different methods.}
	\label{fig:roc_froc_curves} \vspace{-3mm}
\end{figure}

\tableautorefname ~\ref{tab:existing_method_comparison} summarizes the quantitative results of all compared methods and the proposed method. On the more challenging rib fracture detection task, Mean Teacher is the most competitive baseline method, measuring an AUROC of 0.9155 and an FROC score of 0.8540. Our proposed method measures an AUROC of 0.9318 and an FROC score of 0.8914, which significantly outperforms Mean Teacher by a $1.63\%$ gap on the AUROC, and a $3.74\%$ gap on the FROC score. The ROC and FROC curves of the evaluated methods are shown in Fig.~\ref{fig:roc_froc_curves}. On the easier clavicle fracture detection task, CheXNet and Li \etal{}~\cite{li2018thoracic} report the highest AUROCs (\ie, above 0.95) among the baseline methods. Mean Teacher delivers the strongest FROC score of 0.8884 among the baseline methods. Our proposed method also outperforms all baseline methods on the clavicle fracture detection task, reporting an AUROC of 0.9646 and an FROC of 0.9265.

We note that the three knowledge distillation methods, $\rm{\Pi}$-Model, Temporal Ensemble and Mean Teacher, perform stronger than the supervised detection methods. The advantage is more significant on the easier clavicle fracture detection task. This is mainly because clavicle fractures have simpler geometric property and similar visual patterns, which knowledge distillation methods can effectively learn from the pseudo GT of unlabeled data. However, on the more complex rib fracture detection, the advantage of knowledge distillation methods is much less significant. Due to the complex visual patterns of rib fracture and the limited region-labeled positive data, the pseudo GT maps have a low sensitivity (\ie, the supervised pre-trained model reports a low FROC score of 0.7267), which limits the knowledge transferred to the distilled model. Using the proposed \ac{AALS}, our method effectively transfers more knowledge to the student model, hence achieving significantly improved performance compared to the previous knowledge distillation methods. 

We observed that CheXNet and Li \etal{}~\cite{li2018thoracic} significantly outperform baseline knowledge distillation methods on the clavicle fracture AUROC metric, but no performance advantage is observed on the rib fracture AUROC. This is because CheXNet and Li \etal{}~\cite{li2018thoracic} specifically use the positive image-level label, while the baseline knowledge distillation methods do not. In particular, CheXNet is trained via weakly-supervised learning purely using image-level labels. Li \etal{}~\cite{li2018thoracic} exploits image-level positive labels in a multi-instance learning manner. In contrast, the baseline knowledge distillation methods treat the image-level positive images as unlabeled data. While weakly-supervised learning and multi-instance learning are effective on learning the simpler clavicle fractures, they are less effective on more complex rib fractures. In addition, CheXNet and Li \etal{}~\cite{li2018thoracic} also produce poor localization performances. CheXNet provides localization visualization via class activation maps (CAM). Since the CAM values are not comparable across images, the FROC cannot be calculated for CheXNet results. As Li \etal{}~\cite{li2018thoracic} consistently produces overly large activation areas, it does not report meaningful FROC scores. For both CheXNet and Li \etal{}~\cite{li2018thoracic}, we qualitatively verified that their localization performances are worse than other methods, as demonstrated by the examples shown in Fig.~\ref{fig:fracture_visualization}.

\subsection{Ablation Study}
We validate our proposed \ac{AALS} by conducting experiments with different sharpening strengths $a_0$ and centers $t$, respectively. First, to analyze the effect of the label sharpening center $t$, we evaluate \ac{AALS} with $t=0.2,0.3,0.4,0.5$ and summarize the results in Table~\ref{tab:polarize_threshold}. Using $t=0.4$ achieves the best detection performance, measuring the highest/second highest \ac{AUROC} score of $0.9318$/$0.9646$, and the highest \ac{FROC} score of $0.8914$/$0.9265$, on rib/clavicle fracture detection. Note that for clavicle fracture classification, the best \ac{AUROC} score of $0.9661$ achieved at $t=0.2$ is only marginally better than that of $t=0.4$. The sharpening center behaves as a trade-off between sensitivity and specificity. We note that our method consistently outperforms baseline methods using all four $t$ values.
Second, we fix the center $t=0.4$ and evaluate $a_0=1,4,8,16$ to study the impact of the sharpening strength. As summarized in Table~\ref{tab:polarize_strength}, label sharpening with strength $a_0=4$ results in the best detection performance. For $a_0=1$, no label sharpening is applied, which results in degraded performance. For $a_0=8,16$, the label sharpening becomes overly aggressive (as shown in Fig.~\ref{fig:polarization}), which also causes false positives in sharpened pseudo \ac{GT} and hence slight performance degradation.

We further conduct an experiment to study the involvement of image-level positive set $\mathcal{P}$. \figureautorefname~\ref{fig:data_ratio} shows the classification and detection performances for rib and clavicle using a subset of $\mathcal{P}$ with different ratios (0\%, 20\%, 60\%, 100\%), where 0\% and 100\% correspond to the supervised pre-training student model and the proposed method, respectively. We observe that larger $\mathcal{P}$ improves both the classification AUROC and detection FROC scores. This verifies the motivation of our method that CAD model training can benefit from utilizing image-level labels from clinical diagnoses. It also suggests a potential of our method to further improve its performance by incorporating more data with clinical diagnoses without additional annotation efforts. 

\setlength{\tabcolsep}{3pt}
\begin{table}[t]
    \begin{minipage}{0.5\linewidth}
        \centering
        \caption{Study of the sharpening bias.}
        \vspace{1em}
        \begin{tabular}{lcccc}
        \toprule
        \multirow{2}{*}{$t$}  & \multicolumn{2}{c}{Rib fracture} & \multicolumn{2}{c}{Clavicle fracture} \\ 
                                      \cmidrule(lr){2-3}                 \cmidrule(lr){4-5}
                                & AUROC              & FROC             & AUROC              & FROC  \\ \midrule
            0.2                 & 0.9289           & 0.8902           & \textbf{0.9661}  & 0.9236 \\
            0.3                 & 0.9261           & 0.8888           & 0.9611           & 0.9168 \\
            0.4                 & \textbf{0.9318}  & \textbf{0.8914}  & 0.9646           & \textbf{0.9265} \\
            0.5                 & 0.9271           & 0.8848           & 0.9577           & 0.9106 \\
            \bottomrule
        \end{tabular}
        \label{tab:polarize_threshold}
    \end{minipage}
    \begin{minipage}{0.5\linewidth}  
    \centering
    \caption{Study of the sharpening strength.}
    \vspace{1em}
    \begin{tabular}{lcccc}
        \toprule
        \multirow{2}{*}{$a_0$}  & \multicolumn{2}{c}{Rib fracture} & \multicolumn{2}{c}{Clavicle fracture} \\ 
                           \cmidrule(lr){2-3}                 \cmidrule(lr){4-5}
                          & AUROC              & FROC                  & AUROC & FROC  \\ \midrule
        1                 & 0.9222           & 0.8783           & 0.9550             & 0.9036 \\
        4                 & \textbf{0.9318}  & \textbf{0.8914}  & \textbf{0.9646}    & \textbf{0.9265} \\
        8                 & 0.9283           & 0.8884           & 0.9606             & 0.9090 \\
        16                & 0.9302           & 0.8911           & 0.9620             & 0.9185 \\
    \bottomrule
    \end{tabular}
    \label{tab:polarize_strength}
    \end{minipage}
\end{table}

\section{Conclusion}
In this paper, we introduced a specifically-designed \ac{SSL} method to exploit both limited expert annotated region-level labels and large-scale image-level labels mined from the clinical diagnoses records for training a fracture detection model on CXR. We demonstrated that by accounting for the imbalanced data distribution and exploiting the clinical diagnoses, the proposed \ac{AALS} scheme can effectively improve the effectiveness of knowledge distillation on only image-level labeled data. On a large-scale real-world scenario dataset, our method reports the state-of-the-art performance and outperforms previous methods by substantial margins. Our method offers a promising solution to exploit potentially unlimited and automatically mined clinical diagnosis data to facilitate CAD model training.


%
%
%
\bibliographystyle{splncs04}
\bibliography{reference}

\end{document}